# FloorPlan-DeepSeek (FPDS)

## A multimodal approach to floorplan generation using vector-based next room prediction

*Jun Yin[1], Pengyu Zeng[2], Jing Zhong[3], Peilin Li[4], Miao Zhang[5], Ran Luo[6], Shuai Lu[7]*
*[1,2,3,5,7]Tsinghua University, [4]National University of Singapore,*
*[6]South China University of Technology*
*[1,2,3]{yinj24|zeng-py24|zhongj24}@mails.tsinghua.edu.cn*
*[4]e1351227@u.nus.edu [6]201930093463@mail.scut.edu.cn*
*[5,7]{zhangmiao|shuai.lu}@sz.tsinghua.edu.cn*

*In the architectural design process, floor plan generation is inherently progressive and iterative. Architects typically construct floor plans step by step, guided by functional requirements and design intentions, while continuously making adjustments and refinements throughout the design cycle. However, existing generative models for floor plans are predominantly end-to-end generation that produce an entire pixel-based layout in a single pass. This paradigm is often incompatible with the incremental workflows observed in real-world architectural practice. Therefore, there is an urgent need to develop a modeling framework that supports progressive generation, more closely aligning with the actual design process of “constructing the whole from the parts.”*

*To address this issue, we draw inspiration from the autoregressive “next token prediction” mechanism commonly used in large language models, and propose a novel “next room prediction” paradigm tailored to architectural floor plan modeling. The contributions of this study are twofold: (1) We introduce FloorPlan-DeepSeek (FPDS), a floor plan reconstruction method based on the DeepSeek framework, specifically optimized for the vector-based characteristics of architectural images. By representing complete floor plans through the vector data of rooms with distinct functions, this approach enables seamless integration into large language model training pipelines. (2) We conduct extensive experiments across multiple datasets and compare our model against existing baselines. Experimental evaluation indicates that the FPDS model attains an FID of 16.465, PSNR of 76.564, and SSIM of 0.934, indicating its notable performance in generating high-quality complete floor plans. FPDS demonstrates competitive performance in comparison to diffusion models and Tell2Design in the text-to-floorplan task, indicating its potential applicability in supporting future intelligent architectural design*

## INTRODUCTION

Building on recent breakthroughs in machine learning, researchers have explored a growing spectrum of applications in sustainable architecture (Gao et al., 2023; Huang et al., 2025; Li et al., 2019; Zeng et al., 2025; Zhang et al., 2024), energy-conscious design (Chen et al., 2024; Jia et al., n.d.; Lu et al., 2024; Lu et al., 2016; Ma et al., 2025), multimodal content generation (Sun et al., 2025a; Sun et al., 2025b; Zou et al., 2023), visual quality enhancement (Yan et al., 2016; Yin et al., n.d.; Zou et al., 2021), performance-oriented optimization (Lu et al., 2016; Lu et al., 2024; Ma et al., 2025; Yin et al., 2024), and human-AI collaborative systems (Gao et al., 2023; Li et al., 2019; Wang et al., 2024; Yin et al., 2025; Zeng et al., 2025; Zhang et al., 2024).

These developments have also paved the way for rethinking fundamental architectural tasks, including floor plan generation, through data-driven and generative approaches.

As a fundamental component of architectural design, floor plan creation is a dynamic and continuously evolving process. Designers typically rely on vector-based drafting tools such as AutoCAD and Revit to incrementally construct different parts of a building’s floor plan, making ongoing modifications and refinements throughout the design process. However, current AI-driven generative models for floor plan synthesis—such as HouseGAN (Nauata et al., 2020), PlanIT (Wang et al., 2019), and Tell2Design (Leng et al., 2023)—are predominantly based on end-to-end generation strategies. These models take as input various forms of structural or semantic information, including building outlines, functional requirements, or graph-based representations, and directly output a complete pixel-based floor plan. While these methods have achieved notable progress in automating architectural layout design, they still face two critical challenges in terms of practical application.

First, the prevailing end-to-end “one-shot” generation paradigm adopted by current mainstream models lacks interactivity and flexibility. Although it offers efficiency advantages in generating initial design schemes, the generation process remains a “black box,” with limited transparency in intermediate construction stages. Single-pass generation restricts real-time user input and lacks support for progressive design, where structural frameworks precede detailed refinement. In contrast, architectural practice involves iterative development with evaluations across zoning, circulation, and spatial proportioning (Lawson, n.d.). Therefore, constructing a multi-round generation mechanism that supports staged outputs and user-interaction feedback is a critical pathway toward developing generative models into intelligent design-assistance systems.

Second, current models often produce raster images or low-semantic geometric blocks lacking geometric and topological data, limiting their utility in tasks like annotation, component recognition, and semantic interaction. As a result, manual reconstruction is frequently required in design software, reducing efficiency.

To address the aforementioned challenges, emerging autoregressive generation methods in the field of computer vision offer valuable insights. In recent years, the rapid development of large language models (LLMs) has demonstrated exceptional capabilities in tasks involving step-by-step generation. In particular, autoregressive methods based on “next token prediction” have achieved remarkable success in natural language generation tasks (Li et al., 2024). The progressive generation mechanism inherent in LLMs—realized through “next token prediction”—bears a deep resemblance to the iterative thinking process of architectural design. Drawing upon this insight, this study innovatively transfers the paradigm to the domain of architectural floor plan generation by proposing a “next room prediction” approach as shown in figure 1, enabling progressively constructed spatial layouts guided by semantic intentions.

Figure 1
Graphic Abstract

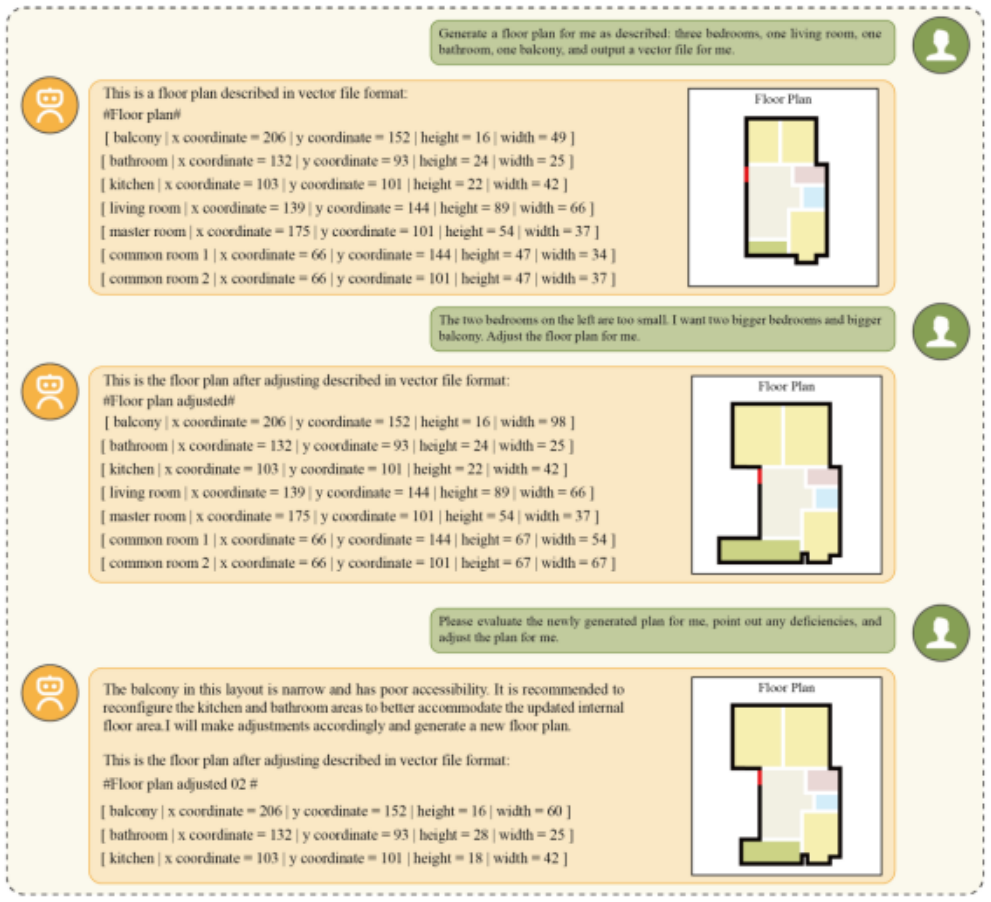

Compared to existing models, our approach offers three key advantages: (1) By adopting an autoregressive generation strategy similar to that of large language models, the system incrementally constructs spatial layouts in a structured sequence, thereby enhancing controllability and interpretability throughout the design process; (2) It supports real-time, interactive modifications based on user feedback, strengthening human-AI collaborative design capabilities and improving both design efficiency and user satisfaction; (3) The output is formatted as standardized vector data, allowing seamless integration into downstream stages of architectural and engineering workflows. Overall, FPDS presents an efficient and promising approach for floor plan generation. It demonstrates the potential to bridge generative AI with real-world architectural design processes, and opens up new directions for intelligent computer-aided design in architecture.

## RELATED WORK

.This study combines recent advances in floor plan generation, large language models (LLMs), and text-to-image synthesis, proposing a vector-based method within the DeepSeek framework. It introduces a "next room prediction" autoregressive strategy, adapts LLMs for architectural vector tasks, and leverages cross-modal alignment techniques to enhance structured generation. The following section reviews related work in these domains

### Floor plan generation

Floor plan generation is a critical topic in computer-aided architectural design. Traditional rule-based methods offer a high degree of controllability but lack flexibility, making them inadequate for addressing complex design requirements (Ruiz-Montiel et al., 2013). With the advancement of deep learning, researchers have proposed generative models based on GANs and VAEs—such as ActFloor-GAN and HouseGAN—which learn from data distributions to generate diverse and functionally reasonable layouts. However, these models primarily produce rasterized images, which lack geometric control and editability. Diffusion models have recently shown remarkable capabilities in image synthesis and have begun to find applications in the generation of architectural visual content. For example, HouseDiffusion integrates discrete and continuous denoising processes to produce layout-consistent workload vector floor plans (Shabani et al., 2023). Nonetheless, such end-to-end approaches still struggle to support interactive and iterative adjustments during the design process.

To improve controllability, studies like PlanIT and Tell2Design adopt modular or graph-based methods that incrementally generate rooms. While structurally expressive, they face limitations in layout control. A room-by-room autoregressive paradigm offers a flexible and interactive alternative.

### Large language models

Large language models (LLMs) have recently driven significant progress in natural language processing, exemplified by notable systems such as the GPT series (Achiam et al., 2023) and DeepSeek (Liu et al., 2024). Autoregressive models such as GPT, based on the “next token prediction” mechanism, have demonstrated powerful contextual modeling and generative consistency in tasks involving text and code. Recent studies have further shown that this generative paradigm can be extended to the visual domain, enabling semantically coherent image and layout generation.

The core advantage of LLMs lies in their capability for contextual understanding and structured output, offering a promising pathway for floor plan generation in architecture. However, architectural graphic data is highly structured, requiring task-specific adaptations and architectural optimizations to enable effective representation and generation of vector-based architectural elements.

## Text-to-image models

Text-to-image models generate visuals aligned with language inputs. Early GAN-based methods used multi-stage attention for detail refinement, while recent diffusion models like DALL • E 2 and Stable Diffusion dominate with high-fidelity synthesis. Transformer architectures now play a central role due to their scalability and sequence modeling power. Building on this foundation, the multimodal model CLIP aligns textual and visual information by learning a joint embedding space (Radford et al., 2021). DALL • E further integrates Transformer and CLIP architectures, significantly improving text-image alignment and generative quality (Ramesh et al., 2022).

Techniques from text-to-image models — especially for semantic alignment and image fidelity — inform floor plan generation, which similarly requires semantic-output consistency. Unlike natural images, floor plans are highly structured, requiring geometric constraints. Combining LLMs with multimodal learning offers a promising path toward vector-based, automated architectural design.

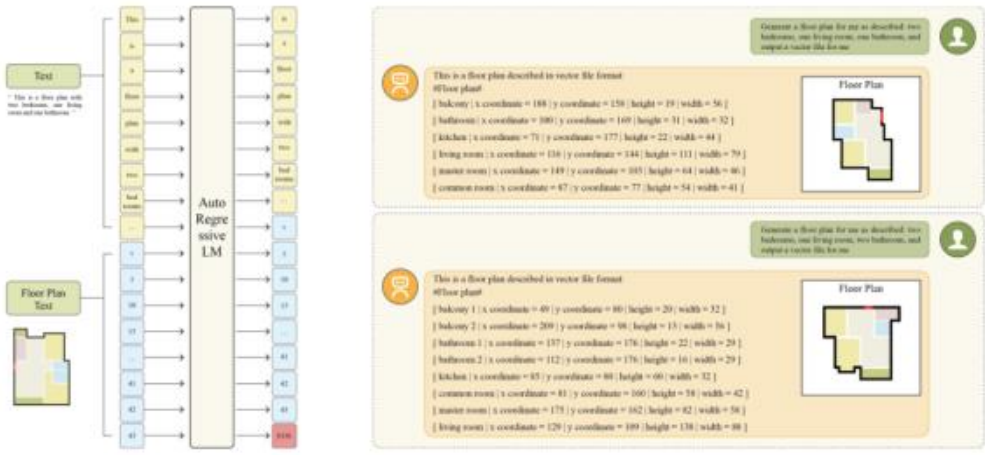

Figure 2
Over view of our method and floor plan generation capability of FPDS

## METHODOLOGY

Inspired by the "text-guided floor plan generation" approach proposed in Tell2Design (Leng et al., 2023), we present a novel method—FloorPlan-DeepSeek (FPDS). The core idea is to transfer the autoregressive mechanism of “next token prediction” from large language models (LLMs) to the task of architectural floor plan generation, introducing a new paradigm of “next room prediction” as shown in figure 2. Similar to Tell2Design, we employ a vectorized representation to describe the properties of individual rooms. However, we improve upon the model architecture and the generation formulation to better accommodate inference within larger-scale language models.

Conditioned on a partially known floor plan layout, the model sequentially predicts the type and spatial attributes of the next room, gradually constructing a complete floor plan. The entire generation process is formalized as a sequence of room vectors, enabling an autoregressive reasoning and construction approach analogous to that used in natural language sequence modeling.

## Data encoding and vector representation

To enable the autoregressive spatial construction process of the model, each room in the architectural floor plan is structurally encoded as a set of vectors, forming a sequential representation of rooms. Similar to the approach adopted in Tell2Design, we describe each room using elements such as position and size. Specifically, $t_i$ denotes the room type (e.g., living room, kitchen, bedroom), which is represented directly using textual labels. $(x_i, y_i)$ refers to the coordinates of the room's center within the floor plan coordinate system, while $(w_i, h_i)$ represents the room ' s span along the horizontal and vertical axes:

$$r_i = (t_i, x_i, y_i, w_i, h_i) \tag{1}$$

As a result, the vector representation for all rooms is organized into a sequence according to the prediction order, as follows:

$$\mathcal{S} = [r_1, r_2, \cdots, r_N] \tag{2}$$

Here, N denotes the total number of rooms in the floor plan. In this study, the room sequence, along with either textual prompts or known partial floor plan features, is fed into a large language model for autoregressive decoding. As illustrated in figure 3, the generation process depicts the complete construction pipeline—from textual input and room vector output to final floor plan synthesis.

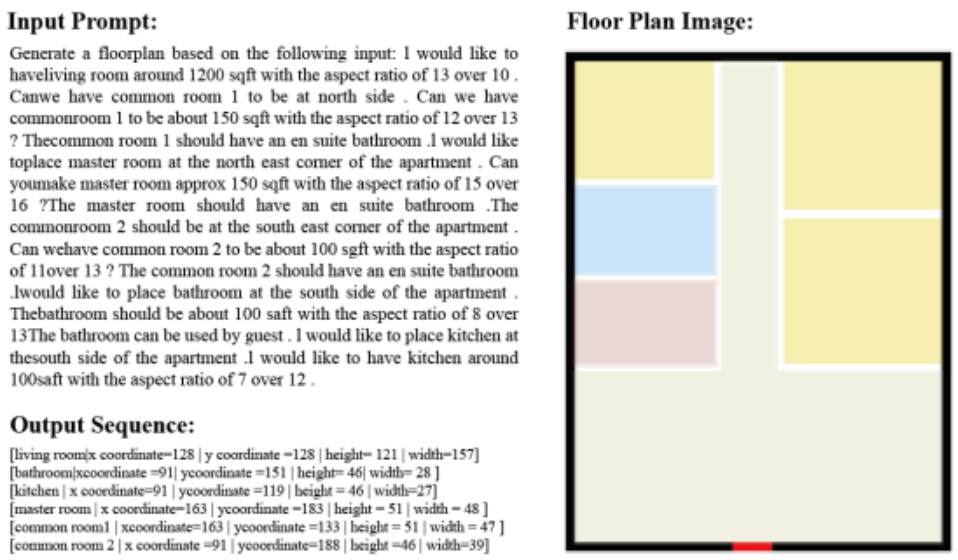

Figure 3 Floorplan generation based on natural language prompt

## Autoregressive generation model

During the training phase, we fine-tune a pretrained large language model to learn the conditional distribution of the next room vector, given an already generated sequence of room vectors and a textual prompt. This process is formalized as a conditional autoregressive generation task, with the objective of maximizing the following conditional likelihood function:

$$P(\mathcal{S} \mid x, \theta) = \prod_{i=1}^{N} P(r_i \mid r_1, \dots, r_{i-1}, x, \theta) \tag{3}$$

$\mathbf{x}$: textual description and/or known partial floor plan information; $\theta$: model parameters.

Compared to Tell2Design, the FPDS model introduces several explicit structural encodings—such as the positional relationships between room types and their coordinates—within its internal representation. By leveraging the multi-head attention mechanism of large language models, FPDS effectively captures contextual dependencies among rooms, thereby enhancing its ability to model spatial logic and functional adjacency.

## Auxiliary constraints and optimization objective

Given the strong structural constraints and functional logic inherent in architectural floor plans, we introduce the following auxiliary constraints into the autoregressive decoding process to ensure the plausibility and editability of the generated results:

Functional Adjacency Constraint: Rooms with spatial dependencies (e.g., bedroom and bathroom) are generated close together, following standard architectural principles.

Editability Constraint: To ensure real-world maintainability, the model generates room boundaries with high local reconstructability, enabling easier post-editing and interaction.

Outline Boundary Constraint: The generated rooms are restricted from extending beyond predefined building outline boundaries, thereby preserving geometric validity and compositional consistency.

To incorporate these constraints into training, we define the total loss as a weighted sum of the negative log-likelihood (NLL) loss from the language modeling objective and the additional constraint losses. Let $\mathcal{L}_{\text{nll}}$ denote the primary language modeling loss and $\mathcal{L}_{\text{cnst}}$ denote the total auxiliary constraint loss. The overall optimization objective is given by:

$$\mathcal{L}(\theta) = \mathcal{L}_{\text{nll}}(\theta) + \lambda\mathcal{L}_{\text{cnst}}(\theta) \tag{4}$$

λ: Balancing Coefficient.This loss imposes a regularization force on the model parameters θ during backpropagation, ensuring the geometric and functional plausibility of the generated room sequences.

## Inference strategy

During inference, FPDS uses autoregressive decoding to generate each room based on prior outputs and textual input. To enhance quality and diversity, we incorporate the following two widely used strategies into the decoding process:

Top-K Sampling: At each prediction step, the model samples from the top K candidates with the highest probabilities, enabling controlled diversity while preventing mode collapse.

Beam Search: For tasks demanding high spatial accuracy, beam search retains top decoding paths to ensure output optimality. The model sequentially generates room vectors, which are assembled into a full vectorized floor plan.

## EXPERIMENT

Figure 4 compares floor plan generation results under various apartment types using Obj-GAN, Tell2Design, Glm-4-9b-chat, Qwen2.5-VL-72B-Instruct, and our FloorPlan-DeepSeek-R1-32B. Each row corresponds to a ground-truth plan from one- to three-bedroom units. Inputs follow the structured format in Figure 3 but are omitted due to length..

The comparison reveals that FloorPlan-DeepSeek consistently achieves better reconstruction of room proportions and spatial-semantic logic across most examples. The generated layouts are not only more structurally coherent but also demonstrate adjacency relationships that align more closely with common architectural design conventions. For instance, in the third row, FloorPlan-DeepSeek successfully produces clearly partitioned functional zones with well-placed kitchen and bathroom areas—outperforming Obj-GAN and GLM4, which show misplaced or disproportionate rooms. Even in more complex three-bedroom configurations (e.g., the last two rows), FloorPlan-DeepSeek maintains effective spatial organization and coherent composition, demonstrating the robustness of the autoregressive paradigm in handling complex spatial semantics.

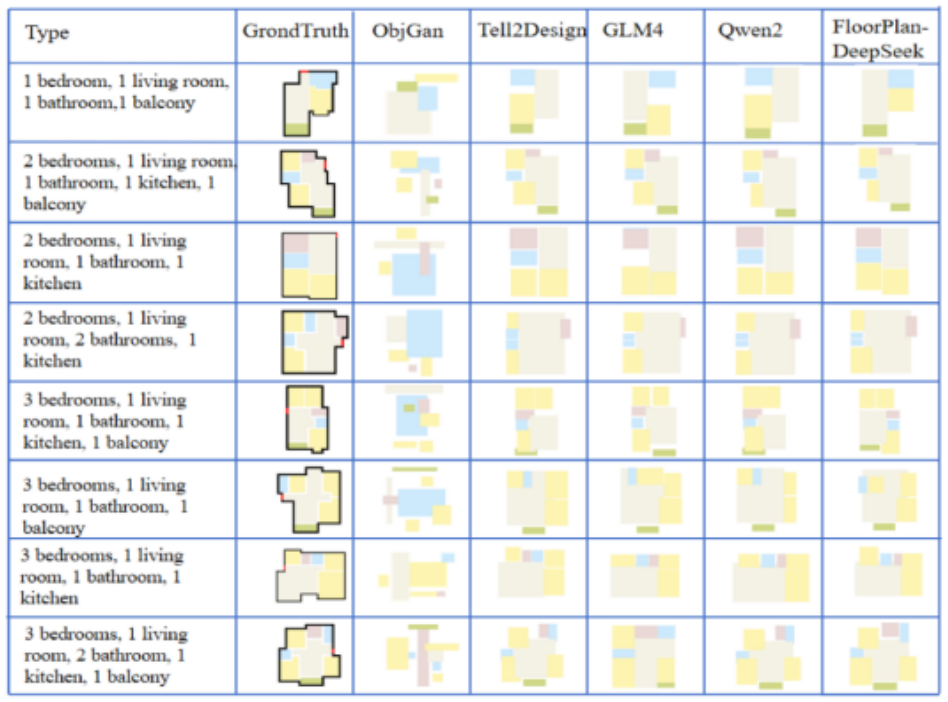

Figure 4 Comparison of floor plan generation results by different models under typical apartment layout conditions

Notably, some models suffer from issues such as unclear room boundaries, disorganized zoning, or incomplete structures. Obj-GAN, in particular, frequently produces overlapping rooms or chaotic layouts. Tell2Design, which serves as a foundational method for this task, performs relatively well in structural layout generation but still falls short in spatial proportion control and functional adjacency. In contrast, FPDS (FloorPlan-DeepSeek) further improves compositional coherence and semantic consistency, reliably generating well-structured and clearly partitioned plans across diverse layout types. The visualization results not only validate FPDS' s superiority in semantic alignment and geometric construction but also highlight its strong generalization capabilities and structural interpretability, offering a more reliable technical pathway for automated architectural layout generation systems.

To evaluate the performance of FloorPlan-DeepSeek (FPDS), we apply three standard image quality metrics: FID, PSNR, and SSIM. Across multiple datasets, FPDS achieves an FID of 16.465, PSNR of 76.564, and SSIM of 0.934, suggesting competitive quality in floor plan generation.FPDS shows strong performance in distributional alignment, reconstruction accuracy, and structural consistency.

## DISCUSSION

In this work, we propose FloorPlan-DeepSeek to explore how large language models can assist floor plan generation. The model demonstrates promising results from natural language inputs, suggesting the potential for a semantics-driven design paradigm. Nonetheless, several practical limitations remain and are discussed below:

(1) Multimodal Robustness: While DeepSeek handles text and image inputs, it struggles with layout errors under complex semantics. Future work should improve model design and fusion strategies to enhance stability.

(2) Generation Accuracy & User Interaction: Despite acceptable spatial accuracy, local mismatches remain. Integrating user feedback and fine-grained constraints (e.g., room spacing, door/window placement) could improve alignment with design intent and semantic precision.

(3) Model Scalability: While effective for single-unit layouts, future work should explore hierarchical generation and semantic cues (e.g., zoning, circulation) to support complex cases like multi-level or mixed-use floor plans.

## CONCLUSION

This paper introduces FloorPlan-DeepSeek, which reformulates "next token prediction" into a "next room prediction" framework for floor plan generation. It outperforms diffusion models and Tell2Design in layout accuracy, consistency, and semantic alignment, while enabling interactive refinement. Future work will enhance multimodal robustness, user interaction, and model scalability.. Overall, FloorPlan-DeepSeek offers a scalable and promising solution that contributes to the broader integration of language models into architectural design workflows

## AUTHORSHIP INFORMATION

Author Contributions: Conceptualization, J.Y.*, P.Z.*, and J.Z.†; methodology, J.Y.*, P.Z.*, and J.Z.†; software, P.L., H.Z., and M.Z.; validation, P.L., T.Z., and J.L.; formal analysis, J.L. and S.L.; investigation, J.Y.* and P.Z.*; data curation, J.L. and M.Z.; writing—original draft preparation, J.Y.* and P.Z.*; writing — review and editing, P.L. and J.Z. †; visualization, P.L. and H.Z.; supervision, J.Z. †; discussion, P.L., T.Z., and M.Z. All authors have read and agreed to the published version of the manuscript.

*These authors contributed equally to this work.

†correspond author.